\documentclass[sigconf]{acmart}
\usepackage{multirow} 
\usepackage{makecell}
\usepackage{tabularx}
\usepackage{booktabs}

\usepackage{array}
\usepackage[caption=false,font=normalsize,labelfont=sf,textfont=sf]{subfig}
\usepackage{textcomp}
\usepackage{stfloats}
\usepackage{url}
\usepackage{verbatim}
\usepackage{graphicx}
\usepackage{epstopdf}
\usepackage{amsmath,amsfonts}
\usepackage{enumitem}
\usepackage{algorithmic}
\usepackage{pifont}
\usepackage{balance} 

\copyrightyear{2025}
\acmYear{2025}
\begin{document}

\title{Lightweight Remote Sensing Scene Classification on Edge Devices via Knowledge Distillation and Early-exit}

\author{Yang Zhao}
\affiliation{%
  \institution{\normalsize{Harbin Institute of Technology, Shenzhen}}
  \city{Shenzhen}
  \state{Guangdong}
  \country{China}
}
\email{yang.zhao@hit.edu.cn}

\author{Shusheng Li}
\affiliation{%
  \institution{\normalsize{Harbin Institute of Technology, Shenzhen}}
  \city{Shenzhen}
  \state{Guangdong}
  \country{China}
}
\email{lss11@foxmail.com}

\author{Xueshang Feng}
\affiliation{%
  \institution{\normalsize{Harbin Institute of Technology, Shenzhen}}
  \city{Shenzhen}
  \state{Guangdong}
  \country{China}
}
\email{fengxueshang@hit.edu.cn}

\renewcommand{\shortauthors}{Zhao et al.}

\begin{abstract}
As the development of lightweight deep learning algorithms, various deep neural network (DNN) models have been proposed for the remote sensing scene classification (RSSC) application. However, it is still challenging for these RSSC models to achieve optimal performance among model accuracy, inference latency, and energy consumption on resource-constrained edge devices. In this paper, we propose a lightweight RSSC framework, which includes a distilled global filter network (GFNet) model and an early-exit mechanism designed for edge devices to achieve state-of-the-art performance. Specifically, we first apply frequency domain distillation on the GFNet model to reduce model size. Then we design a dynamic early-exit model tailored for DNN models on edge devices to further improve model inference efficiency. We evaluate our E3C model on three edge devices across four datasets. Extensive experimental results show that it achieves an average of 1.3x speedup on model inference and over 40\% improvement on energy efficiency, while maintaining high classification accuracy. 
\end{abstract}



\maketitle

\pagestyle{plain} 
    
\section{Introduction} \label{S:introduction}

As the development of deep neural network (DNN) and embedded system techniques, remote sensing scene classification (RSSC) on edge devices becomes more and more practical for land resource management, urban planning, traffic flow prediction and many other remote sensing applications~\cite{cheng2017remote, 9619948, wu2024takd}.
For traditional RSSC, remote sensing imagery data are first downloaded from the sensor sources, such as RGB-infrared cameras on unmanned aerial vehicles (UAVs) or hyperspectral sensors on satellites. Then they are pre-processed and fed into DNN models, which are performed on high-performance servers~\cite{cheng2017remote}.
However, the resolution and volume of remote sensing imagery data have been rapidly increasing as the development of sensor technology. It will create overwhelming pressure on data communication systems if large amounts of imagery data are transferred to a central server, especially for real-time applications~\cite{cheng2020remote,wang2024luojia,jin2025fed}.
Thus, it becomes a promising solution to perform DNN model inference on edge devices near data sources, without suffering from the communication bottleneck issue~\cite{giuffrida2021varphi,xu2023ai,jiang2025efficient}.

\begin{figure*}[htbp]
\centering
\includegraphics[width=1\linewidth]{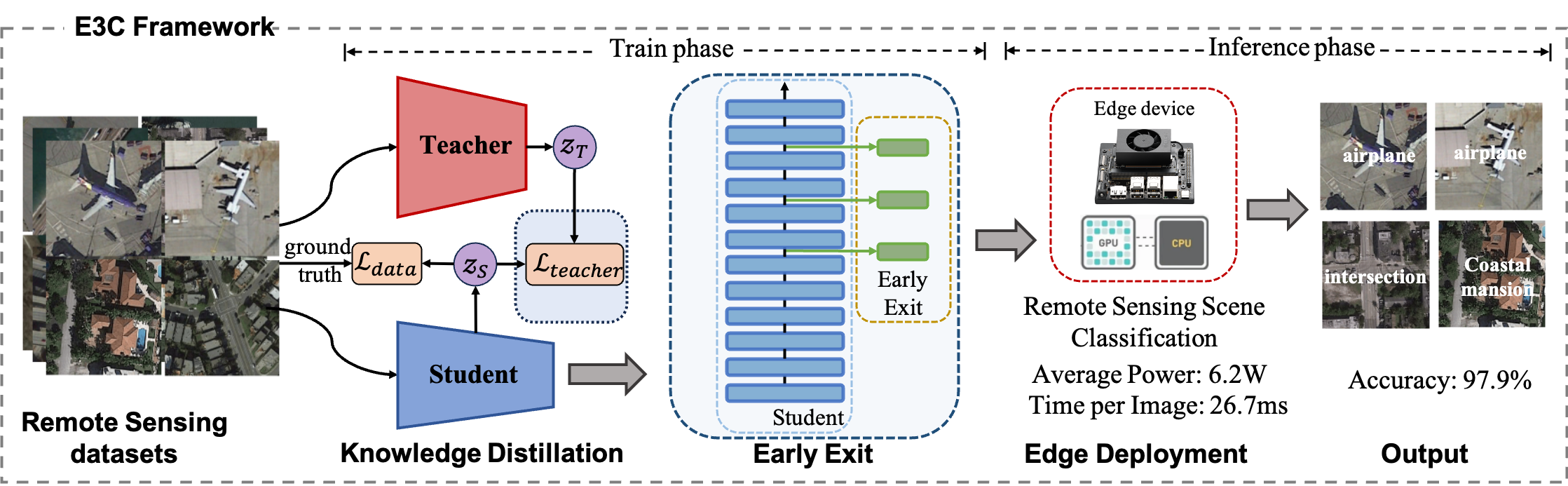}
\caption{Overview of the E3C Framework with knowledge distillation and early-exit tailored for RSSC on edge devices. }
\label{fig:overview}
\end{figure*}

However, edge devices usually has limited resources, and embedded systems such as UAVs have strict requirements on size, weight and power consumption~\cite{shuvo2022efficient}. Although vision Transformers (ViT) and other DNN models can provide outstanding classification accuracy for RSSC~\cite{yang2023explainable}, these models require substantial model parameters and floating-point operations (FLOPs)~\cite{yang2023sagn}, and thus have high inference latency and energy consumption on edge devices~\cite{wu2024takd, zhang2025e4}. 
To address this challenge, recent studies have developed various lightweight deep learning techniques, such as knowledge distillation, to reduce model complexity and accelerate model inference~\cite{wu2024takd, lu2023energy, zhang2025e4, shuvo2022efficient}. For example, the data-efficient image Transformer (DeiT) model introduces a token-based distillation strategy that significantly outperforms vanilla distillation methods~\cite{touvron2021training}. Early-exiting dynamic neural networks accelerate DNN inference by allowing a model to make predictions from intermediate layers~\cite{regol2023jointly, teerapittayanon2016branchynet}. On the model architecture side, a simple yet computationally efficient architecture called Global Filter Network (GFNet) replaces the self-attention sub-layer in vision Transformers (ViT) with 2D discrete Fourier transform modules and global filter modules that enjoy log-linear computation complexity~\cite{rao2023gfnet}. 
We perform experiments to train these lightweight models, deploy and evaluate them on various edge devices, as shown in Fig.~\ref{fig:overview}. 
However, we find the following issues, when comparing the performance of the GFNet model and vision Transformer models~\cite{touvron2021training, han2022survey}.

First, we find that none of the GFNet, DeiT and ViT models can achieve low latency, e.g., <500 milliseconds, using popular RSSC datasets on resource-constrained edge devices. One example is shown in Fig.~\ref{fig:latency}(c): the GFNet and ViT models exhibit high inference latency, exceeding one second across all three datasets on the resource-constrained Raspberry Pi device. 
We also find that although GFNet is more computationally efficient than ViT, a ViT model combined with early-exit mechanisms achieves even lower latency than GFNet on certain edge devices, since model inference can be completed early by exiting from an intermediate layer of a more sophisticated DNN model. Can we apply lightweight techniques such as knowledge distillation and early-exit to a computationally efficient DNN model other than Transformers to further improve the performance of RSSC on edge devices? We perform the following investigations to find out the answer. 
First, on the knowledge distillation side, we adopt the hard-label distillation approach as in~\cite{touvron2021training} to apply frequency domain distillation on the GFNet model. Specifically, we introduce a distillation token that captures global information by interacting with other tokens through the global filter layer in GFNet~\cite{rao2023gfnet}. We find that the model size and inference latency of the distilled GFNet model can be significantly reduced.

Second, on the early-exit side, we have investigated directly applying the early-exit mechanisms used in JEI-DNN~\cite{regol2023jointly} to the GFNet model to further improve the RSSC performance. However, we find that applying the JET-DNN early-exit method to GFNet does not improve classification accuracy and energy consumption, and even increases latency on certain edge devices. After analyzing our data, it turns out that since we use a relatively lightweight GFNet model as the backbone model, the computational overhead introduced by gating mechanisms (GMs) and intermediate inference modules (IMs) as in \cite{regol2023jointly} outweighs the benefits provided by the early-exit mechanisms, especially when some inputs are not ``easy enough'' to be determined by early classifiers. The JEI-DNN method optimizes a loss that jointly assesses accuracy and inference cost, but it also adds early-exit branches with GMs and IMs modules to each and every layer of the backbone model~\cite{regol2023jointly}, which can cause additional computation overhead. Thus, we propose a new early-exit method tailored for lightweight models on edge devices, in which we start incorporating the early-exit branches from an intermediate layer every other layer of the backbone model. Since the shallow layers of a lightweight model, e.g., GFNet, usually cannot capture image features effectively, significant computational overhead can be saved by using this new method. For edge devices with heterogeneous processors, we propose to use CPUs to compute early-exit model parameters, while loading other model parameters on GPUs, to better utilize heterogeneous resources.

Finally, we design and implement a remote sensing scene classification framework called E3C, which includes knowledge distillation and early-exit mechanisms tailored for the GFNet model on edge devices. We perform extensive evaluations across four RSSC datasets on three different edge devices. Experimental results show that compared with state-of-the-art baselines, E3C achieves an average of 1.3x model inference speedup, and over 40\% energy efficiency improvement, while maintaining high accuracy. 

To summarize, this paper aims to fill current research gaps by investigating lightweight techniques including knowledge distillation and early-exit on a computationally efficient DNN model for the RSSC problem on resource-constrained edge devices.
Through knowledge distillation, the distilled GFNet model achieves high model accuracy with model size and inference latency significantly reduced. A new early-exit model is designed for RSSC model inference on edge devices, which further reduces model latency and energy consumption. The contributions of this paper are as follows:
\begin{itemize}[leftmargin=*]
\item
We investigate lightweight DNN techniques and models for the RSSC problem, and find the root causes for the high inference latency issue on resource-constrained edge devices.
\item
We propose frequency distillation on the GFNet model and a dynamic early-exit model tailored for DNN models on edge devices. We find that knowledge distillation and early-exit mechanisms complement each other in enhancing model efficiency.
\item 
We design a lightweight RSSC framework by integrating proposed DNN models and optimizing the use of heterogeneous computing resources. Extensive evaluation across four datasets on three edge devices show that our E3C model achieves state-of-the-art trade-off performance in terms of accuracy, inference latency and energy efficiency. 
\end{itemize}

\section{Related Work}

\textbf{Remote Sensing Scene Classification.}
Remote sensing scene classification (RSSC) is a fundamental task of interpreting remote sensing imagery data to label them with correct semantic categories. The key to RSSC is to extract informative and discriminative features~\cite{peng2020efficient}.
DNN-based RSSC approaches have the capability of extracting discriminative and semantic features more effectively, compared with traditional handcrafted feature extractors. 
For example, a multiscale spatial-frequency Transformer model uses self-attention to better capture the global dependence of the spatial and frequency multiscale features for RSSC~\cite{yang2023explainable}. 
CGINet proposes a light context-aware attention block to explicitly model the global context to obtain larger receptive fields and more contextual information, so that global context and part-level discriminative features can be combined within a unified framework for accurate RSSC~\cite{zhao2024co}.
The RSSC models mentioned above are accompanied by huge computational costs. Thus, recent studies have investigated various lightweight DNN techniques to improve the efficiency of these RSSC models~\cite{peng2020efficient}. For example, SDT2Net incorporates an efficient attention block and a differentiable token compression mechanism, which adaptively prunes tokens under a given FLOPS constraint to build an efficient RSSC Transformer model~\cite{ni2024remote}. EFPF is an energy-based filter pruning framework designed to reduce model size, which calculates layer energy using the eigenvalues of each weight tensor through singular value decomposition~\cite{lu2023energy}.
While these lightweight DNN models can reduce model size and FLOPS requirements, little work focuses on optimizing and deploying DNN models on resource-constrained edge devices for real-time RSSC model inference.

\textbf{Knowledge Distillation.} Knowledge distillation is an effective model compression technique to improve the efficiency of DNN models without increasing the number of parameters or computational cost. 
The application of knowledge distillation is pioneered by \cite{hinton2015distilling}, and it has becomes a popular technique for transferring knowledge from large teacher models to smaller student models. Nowadays, it is primarily categorized into response-based methods, feature-based methods and relation-based methods~\cite{yang2024survey}. In this paper, we adopt the response-based method, also known as the logit-based method, which involves transferring knowledge through the final classification layer or logits of the model. 
Recently, methods have been proposed to compress remote sensing scene classifiers using knowledge distillation. For example, a target-aware knowledge distillation method introduces a target extraction module, and then a target-aware loss is designed to enable the transfer of knowledge in the target regions from the teacher model~\cite{wu2024takd}. This method can reduce RSSC model size significantly, but it also adds an additional target extraction module, which increases computational overhead. 

\textbf{Early-exit DNN Models.} 
The early-exit DNN models belong to a broader category of dynamic neural networks~\cite{han2021dynamic}, in which the model depth is dynamic by the early-exit mechanism to allowing ``easy'' samples to be output at shallow exits without executing deeper layers. 
The goal of the early-exit mechanism is to augment the backbone model with additional trainable components so as to obtain a final result at a reduced inference cost, and BranchyNet is among the first initiatives to propose adding early-exit branches to DNN models to reduce model inference time~\cite{teerapittayanon2016branchynet}. 
For time-sensitive edge computing applications, various early-exit methods have been proposed to achieve better inference performance~\cite{li2023predictive, liu2023resource}.
For example, TLEE proposes to apply the exit-early mechanism to video recognition, which automatically determines when and where to exit the inference process based on complexity and performance requirements of the input video~\cite{wang2023tlee}. In addition, E4 proposes an energy-efficient DNN inference framework for edge video analysis, which achieves state-of-the-art DNN inference performance by integrating a early-exit module with a DVFS regulator to adjust CPU and GPU clock frequencies optimally based on the DNN exit points~\cite{zhang2025e4}. 
However, we have found little work on integrating early-exit mechanisms with other lightweight techniques for real-time RSSC on edge devices. 
In this paper, our early-exit model is built upon JEI-DNN, which enhances the backbone model with jointly optimized gating and intermediate inference modules~\cite{regol2023jointly}. 

\section{Methodology}
The overview of our lightweight RSSC framework is presented, followed by detailed introduction of two key components. 

\subsection{Overview}
The overview of our E3C Framework is shown in Fig~\ref{fig:overview}, which includes knowledge distillation and \textbf{E}arly-\textbf{E}xit for \textbf{E}dge \textbf{C}omputing-oriented (E3C) RSSC. 
The framework uses a computationally efficient GFNet as the backbone model, which learns long-range spatial dependencies in the frequency domain with a computational complexity of $O(nlogn)$.
In our model training phase, we first train a distilled GFNet model using the logit-based knowledge distillation method in the frequency domain. Then, we train the gating and intermediate inference modules in the early-exit model in an alternating way~\cite{regol2023jointly}.
Finally, we deploy trained models on edge devices to perform model inference. For edge devices with heterogeneous CPU-GPU processors, we separate the compute of model parameters on different processors to better utilize the computational resources. Our E3C model can achieve real-time RSSC inference with high classification accuracy on various edge devices. 

\begin{figure*}
\centering
\includegraphics[width=1\linewidth]{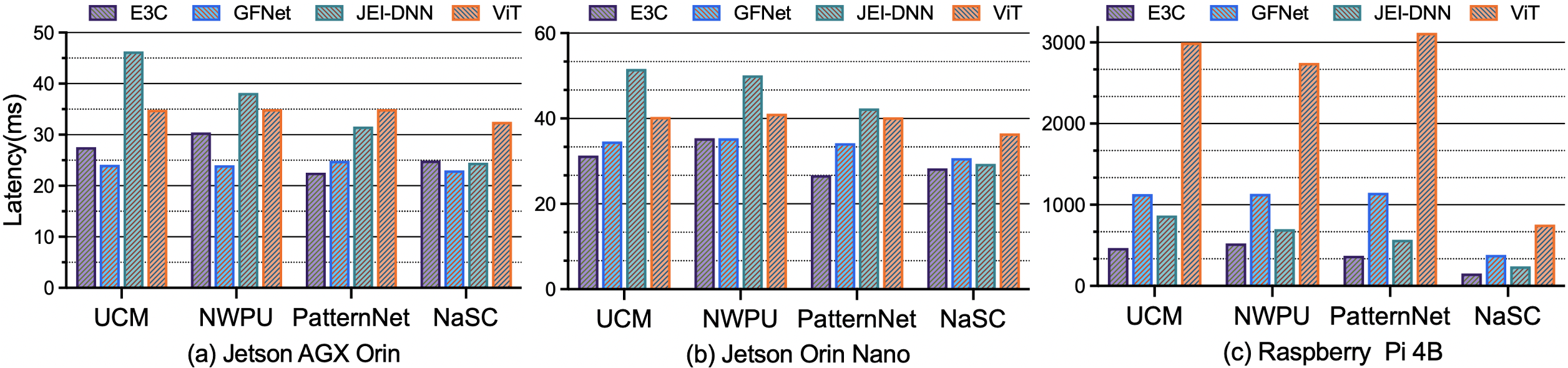}
\caption{Comparison of model inference latency on three edge devices across four RSSC datasets.}
\label{fig:latency}
\end{figure*}

\subsection{Knowledge Distillation on GFNet Model} \label{S:distillation}
Now we introduce the GFNet model and describe how we apply knowledge distillation on it to further improve computation efficiency. 
Given an input image $\mathbf{x} \in \mathbb{R}^{H \times W \times C}$, a 2D discrete Fourier transform $\mathcal{F}[\cdot]$ can be performed to convert the input spatial features to the frequency domain: $\mathbf{X} = \mathcal{F}[\mathbf{x}] \in \mathbb{C}^{H \times W \times C}$. 
Then the spectrum $\mathbf{X}$ in the frequency domain can be modulated by multiplying a learnable global filter \( \mathbf{K} \in \mathbb{C}^{H \times W \times C}\): 
\begin{equation}
    \tilde{\mathbf{X}} = \mathbf{K} \odot \mathbf{X},
\end{equation}
where $\odot$ is the element-wise multiplication, also known as the Hadamard product~\cite{styan1973hadamard}. 
Finally, the modulated spectrum $\tilde{\mathbf{X}}$ can be transformed back to the spatial domain using a 2D inverse Fourier transform.
As shown in \cite{rao2023gfnet}, the global filter $\mathbf{K}$ is equivalent to a depthwise global circular convolution with a complexity of $\mathcal{O}(CS^{2})$, where $S=HW$, and can learn features effectively in the frequency domain. Since the 2D Fourier and inverse Fourier transforms have fast Fourier transforms (FFT) algorithm implementations, and the global filter in the frequency domain enjoys a complexity of $\mathcal{O}(CS\log S)$, GFNet is a more efficient DNN model than ViT models. 

As mentioned in Section~\ref{S:introduction}, to further improve the computational efficiency, we apply knowledge distillation on the GFNet model. Specifically, we use the effective hard-label distillation loss from the Data-efficient image Transformer (DeiT) model~\cite{touvron2021training} as the total loss for our distillation: $\mathcal{L}_{\text{hardDistill}} = \tfrac{1}{2} \mathcal{L}_{CE}^{y} + \tfrac{1}{2} \mathcal{L}_{CE}^{teacher}$, where $\mathcal{L}_{CE}^{y}$ denotes the cross entropy loss from the ground-truth $y$, and $\mathcal{L}_{CE}^{teacher}$ denotes the cross entropy loss from the GFNet teacher model. 
Since GFNet uses FFT to transform inputs into the frequency domain, we propose to perform frequency domain distillation using a distillation token and a class token. The class token can learn the inherent features and distribution of the data~\cite{dosovitskiy2020image}, while the distillation token learns the network information of the teacher model~\cite{touvron2021training}. 
Note that our goal is to investigate the idea of using knowledge distillation methods to further improve the model inference efficiency of the GFNet model on edge devices. We choose the hard-label distillation method, since it is parameter-free and simpler compared with the soft distillation method, but other distillation methods can also be used within our framework.

\subsection{Early-exit Models} \label{S:early-exit}

For the early-exit model, we first introduce a state-of-the-art early-exit dynamic networks called JEI-DNN, then describe how we build upon it to design a new early-exit mechanism for lightweight model inference on edge devices. 

\subsubsection{JEI-DNN Model} \label{S:jei}

Traditional early-exit models add two additional components to the backbone model: 1) inference modules (IMs), also known as intermediate classifiers to generate results at particular network depths, 2) gating mechanisms (GMs) for determining which IMs should be used to derive the final result~\cite{han2021dynamic}. 
However, in most early-exit dynamic networks, the IMs and GMs are decoupled in training, since simple threshold-based gating mechanisms are used with the GMs treated as a post-training add-on component. In addition, for edge computing-oriented settings where DNN models can produce different outputs with varying computational resources, confidence measures are essential but are missing from traditional early-exit models. 
Thus, the state-of-the-art early-exit model JEI-DNN introduces an approach for modeling the probability of exiting at a particular inference module. It also proposes to avoid train-test mismatch and provide good uncertainty characterization by optimizing a loss that jointly assesses accuracy and inference cost~\cite{regol2023jointly}:
\begin{equation} \label{E:jei-loss}
    \mathcal{L}_b=\mathcal{L}_{CE}(p_b,y)+\lambda IC_b,
\end{equation}
where $b$ represents an early exit branch point, i.e., exit at layer \( b \), $p_b$ is the prediction probability, \(\textit{IC}_b\) is the inference cost when the model chooses to exit at layer \( b \), \(\lambda\) is the trade-off coefficient between inference cost and model accuracy representing the importance of inference cost. 

However, directly optimizing (\ref{E:jei-loss}) is challenging, and JEI-DNN adopts an alternating training strategy for updating the parameters of the GMs and IMs. Specifically, four uncertainty statistics are used as the feature inputs for GMs: the maximum predicted probability \( \max(p_b) \), the entropy of the prediction \( \sum_k p_b^k \log(p_b^k) \), the entropy scaled by a temperature \( \sum_k \tilde{p}_b^k \log(\tilde{p}_b^k) \) and the difference between the two most confident predictions~\cite{regol2023jointly}. For IMs, a single layer neural network is used to allow inserting exit branches at every layer. Note that while the GMs and IMs modules are trained alternately, our GFNet-based backbone model is trained separately. That is, the backbone model parameters are frozen, while the parameters of the gate modules and the intermediate classifiers are updating. Also note that the GMs have a gating threshold parameter, whose impact is investigated in the evaluation section.

\subsubsection{Lightweight Early-exit Model} \label{S:lightweight-early-exit}

As mentioned in Section~\ref{S:introduction}, we find that the JEI-DNN model does not always improve DNN inference performance on resource-constrained edge devices. The reason lies in the fact that although the intermediate classifier is a lightweight single neural network, adding IMs and GMs at each and every layer of the backbone model can still increase computational overhead, especially when ``early classifiers'' cannot complete the inference tasks for certain difficult inputs. Thus, we propose a new early-exit model tailored for real-time RSSC on edge devices by improving the JEI-DNN model in the following way.

First, since we use a distilled GFNet model as the backbone of our framework, adding early exit branches to the shallow layers of the distilled model will increase computational redundancy during model inference.
Thus, we propose to start incorporating early-exit branches from an intermediate layer with an index of $l_m$. 
In addition, instead of adding early-exit branches on each and every layer of the model, we propose to incorporate early-exit branches every $M$ layers, so as to reduce computational redundancy while ensuring high accuracy:
\begin{equation}
    \mathcal{B} = \{ b \; ; \; l_m \leq b < L, \; (b - l_m)\% M = 0 \},
\end{equation}
where \( \mathcal{B} \) is the set of the early-exit insertion points, and the total number of GFNet blocks is $L-1$. Note that we investigate the impacts of the parameters $l_m$ and $M$ in our evaluation section. 

Finally, many edge computing devices have heterogeneous processors of CPUs and GPUs. We propose the following mechanism for the early-exit model to better utilize the CPU-GPU heterogeneous computational resources. As mentioned in Section~\ref{S:jei}, uncertainty statistics with logarithmic operations need to be computed for the gating mechanisms of the early-exit model. Thus, we load the computation of the early-exit model parameters on CPUs, while keeping other model parameters on GPUs during model inference. This design mechanism further reduces inference latency by better utilizing heterogeneous processors on edge devices.

\section{Experiments and Results} \label{S:evaluation}
This section presents the E3C implementation details, followed by experimental settings and performance evaluation.

\subsection{Implementation Details} 
As shown in Fig~\ref{fig:overview}, the E3C framework has a training phase and an inference phase. 
In the training phase, the distillation procedures in DeiT~\cite{touvron2021training} are used to perform frequency domain distillation on the GFNet model to train a distilled GFNet model, during which the learning rate is set to be 0.01 and the number of epochs to be 300. 
Then, for training the early-exit model, the distilled GFNet backbone network is frozen, while the IMs and GMs modules are trained as discussed in Section~\ref{S:jei}. We start by training all IMs on the full dataset in a warm-up stage, which ensures that IMs are performing reasonably well when we start training the GMs modules~\cite{regol2023jointly}.
We adopt an alternating training strategy for updating the parameters of the IMs and GMs, during which the learning rate is set to be 0.01 with 200 training iterations. 
Four Nvidia RTX 4090 GPUs are used in the training phase, after which the trained E3C model is deployed on edge devices to perform model inference. Note that DNN model inference optimization techniques, such as TVM~\cite{chen2018tvm}, are orthogonal to the techniques used in our E3C model, and can be used to further improve model performance in future. 

For the early-exit model parameter determination, we use the following statistics from the training data. We define exit rate as the proportion of samples exiting at a particular layer relative to the total number of input samples, and define exit accuracy as the proportion of correctly classified samples among those exiting at that layer. We can derive these statistics during our model training, with examples shown in Fig.~\ref{fig:four_datasets} and our supplementary material. We see that the histograms of the exit rate metric with high exit accuracy starts from the 5th layer of the model for most of the training datasets. Since the JEI-DNN model is proposed to avoid train-test mismatch~\cite{regol2023jointly}, we use the statistics from the training dataset to set the early-exit model parameter $l_m=4$, i.e., use the 5th layer as the starting point of inserting early-exit branches.  
Similarly for the layer interval parameter $M$, we find that the best latency and energy efficiency can be achieved by setting $M = 2$. If setting otherwise, e.g., $M = 3$, adding early-exit branches every three layers, it would produce insufficient exit points, especially considering the fact that we incorporate early exit branches from an intermediate layer. Note that for simplicity, we use the same early-exit configuration parameters $l_m$ and $M$ for all four RSSC datasets in our evaluation. However, the early-exit configurations can be made adaptively based on dataset complexity to further improve the performance, which we leave as future work.

In the model inference phase, we deploy the E3C model combining the strengths of a distilled GFNet model and an early-exit model on three different edge computing devices: NVIDIA Jetson AGX Orin, NVIDIA Jetson Orin Nano and Raspberry Pi 4B. The specifications of these edge devices are listed in Table~\ref{Table:device}. For measuring the power consumption of the edge devices during inference, we use jetson-stats~\cite{jetson-stats} and Power-Z KT002, which is a portable instrument that measures the current and voltage of edge devices. 
From Table~\ref{Table:device}, we see that the Jetson AGX Orin edge device has high-performance CPU-GPU modules with 64GB RAM delivering up to 275 TOPS of compute performance. Jetson Orin Nano also has heterogeneous CPU-GPU modules, but only has 20 TOPS and 8GB RAM, while Raspberry Pi 4B only has 4GB RAM without GPUs, the fewest computational resources among the three edge devices. In Section~\ref{S:evaluation}, we show that our E3C model outperforms existing state-of-the-art models on these edge devices. Python and the PyTorch library are used in our implementation, and our code and data are publicly available at \url{https://github.com/KaHim-Lo/GFNet-Dynn}. 

\begin{table}[t!]
\centering
\caption{Edge devices used in this study.} 
\label{Table:device}
\begin{tabular}{cc} 
\toprule               
          \textbf{Edge Device}& \textbf{Specs.}\\ 
\midrule               
Jetson Orin Nano& \begin{tabular}{@{}l@{}}CPU: 6xCortex-A78AE@1.5GHz \\ GPU: 512xAmpere@0.6GHz \end{tabular} \\
\midrule
Jetson AGX Orin& \begin{tabular}{@{}l@{}}CPU: 12xCortex-A78AE@2.2GHz \\ GPU: 2048xAmpere@1.3GHz \end{tabular} \\
\midrule
Raspberry Pi 4B&  CPU: 4xCortex-A72@1.5GHz  \\ 
\bottomrule            
\end{tabular}
\end{table}

\begin{table*}[ht]
\centering
\caption{Comparison with baselines and ablation studies of distillation and early exit across four datasets on three edge devices.}
\label{tab:ablation}
\begin{tabular}{lcccccccccc}
\toprule
\multirow{3}{*}{Dataset} & \multirow{3}{*}{Distillation} & \multirow{3}{*}{Early Exit}  & \multirow{3}{*}{Params (MB)}  & \multirow{3}{*}{Accuracy (\%)}  &\multicolumn{2}{c}{Jetson AGX Orin}& \multicolumn{2}{c}{Jetson Orin Nano} & \multicolumn{2}{c}{Raspberry Pi 4B} \\
\cmidrule(lr){6-7} \cmidrule(lr){8-9} \cmidrule(lr){10-11}
 &  &  &  &  & \makecell[c]{Energy\\(mJ)}& \makecell[c]{Improve\\(\%)}& \makecell[c]{Energy\\(mJ)} & \makecell[c]{Improve\\(\%)} & \makecell[c]{Energy\\(mJ)} & \makecell[c]{Improve\\(\%)} \\
\midrule
 & \textcolor{red}{\ding{55}}&\textcolor{red}{\ding{55}}& 15.92& 99.3& 144.8& N/A& 96.9& N/A& 1233.5& N/A\\
UCM& \textcolor{green}{\ding{51}}&\textcolor{red}{\ding{55}}& 4.55& 98.6& 75.3& 48.0& 49.8& 48.6& 547.2& 	55.6\\
& \textcolor{red}{\ding{55}}&\textcolor{green}{\ding{51}}& 15.95& 99.3& 126.0& 13.0& 86.3& 10.9& 1105.3& 10.4\\
& \textcolor{green}{\ding{51}}&\textcolor{green}{\ding{51}}& \textbf{4.57}& \textbf{97.9}& \textbf{68.8}& \textbf{52.5}& \textbf{40.7}& \textbf{58.0}	& \textbf{432.2}&	\textbf{65.0}\\
\midrule
   & \textcolor{red}{\ding{55}}&\textcolor{red}{\ding{55}}& 15.93& 96.4& 141.6& N/A	& 99.1& N/A	& 1237.8& N/A	\\
NWPU& \textcolor{green}{\ding{51}}&\textcolor{red}{\ding{55}}& 4.56& 95.0& 73.2& 48.3	& 51.0	& 48.5	& 538.7& 56.4
\\
& \textcolor{red}{\ding{55}}&\textcolor{green}{\ding{51}}& 16.00& 96.3& 108.8& 23.1& 73.1	& 26.2	& 948.0& 23.4\\
 & \textcolor{green}{\ding{51}}&\textcolor{green}{\ding{51}}& \textbf{4.60}& \textbf{94.7}& \textbf{88.2}	& \textbf{41.9}	& \textbf{46.0}	& \textbf{53.6}	& \textbf{480.7}	&\textbf{61.2}
\\
 \midrule
   & \textcolor{red}{\ding{55}}&\textcolor{red}{\ding{55}}& 15.93& 99.7& 144.4	& N/A	& 95.8	& N/A	& 1378.9	& N/A
\\
PatternNet& \textcolor{green}{\ding{51}}&\textcolor{red}{\ding{55}}& 4.56& 98.9& 73.2	& 49.3	& 48.2	& 50.0	& 500.9	& 63.7\\
& \textcolor{red}{\ding{55}}&\textcolor{green}{\ding{51}}& 15.99& 99.5& 101.5	& 29.7	& 68.3	& 28.7	& 903.8& 34.5\\
 & \textcolor{green}{\ding{51}}&\textcolor{green}{\ding{51}}& \textbf{4.59}& \textbf{97.9}& \textbf{58.5}	& \textbf{59.5}	& \textbf{34.7}& \textbf{63.8}	& \textbf{340.6}&\textbf{75.3}
\\
 \midrule
   & \textcolor{red}{\ding{55}}&\textcolor{red}{\ding{55}}& 14.89& 99.1& 71.3& N/A	& 39.9	& N/A	& 412.1& 	N/A\\
NaSC& \textcolor{green}{\ding{51}}&\textcolor{red}{\ding{55}}& 3.92& 98.3& 51.7& 27.5	& 28.6	& 28.3	& 151.8& 63.2
\\
& \textcolor{red}{\ding{55}}&\textcolor{green}{\ding{51}}& 14.90& 98.8& 49.9& 30.0	& 29.4& 26.3	& 256.5& 37.8\\
 & \textcolor{green}{\ding{51}}&\textcolor{green}{\ding{51}}& \textbf{3.93}& \textbf{97.4}& \textbf{49.8}& \textbf{30.2}& \textbf{22.6}	& \textbf{43.3}	& \textbf{140.2}&\textbf{66.0}
\\
 \bottomrule
\end{tabular}
\end{table*}

\subsection{Experimental Settings} \label{S:settings}

\textbf{Datasets.} We evaluate our E3C model using four RSSC datasets: NaSC-TG2~\cite{zhou2021nasc}, PatternNet~\cite{zhou2018patternnet}, UC Merced Land-Use~\cite{yang2010bag}, and NWPU-RESISC45~\cite{cheng2017remote}. 
NaSC-TG2 is an RSSC dataset from the Tiangong-2 space station, which comprises 20,000 remote sensing images, divided into 10 natural scenes, with each scene containing 2,000 images of 128×128 pixels~\cite{zhou2021nasc}. 
PatternNet is a large-scale high-resolution remote sensing dataset collected for remote sensing image retrieval. There are 38 classes and each class has 800 images of size 256×256 pixels~\cite{zhou2018patternnet}. 
The UC Merced (UCM) dataset is one of the most widely used remote sensing benchmarks, which contains 21 scene categories, with 100 images of 256×256 pixels per category~\cite{yang2010bag}. The NWPU-RESISC45 dataset contains 45 scene categories with variations in spatial resolution, object pose, translation, illumination, viewpoint, occlusion, and background~\cite{cheng2017remote}. 
While other aerial scene classification datasets, such as AID~\cite{xia2017aid} can also be used in the evaluation, the dataset diversities mentioned above validate generalizability across common RSSC challenges. 
For example, in terms of diversity in spatial resolution, PatternNet has the highest resolution ranging from 0.062m to 4.693m, while that of NWPU-RESISC45 ranges from 0.2m to 30m. 
Note that to ensure consistent input resolution across all datasets, all images were resized to a uniform resolution of 256×256 pixels in our evaluation. 
We also randomly split each of the datasets into training sets with the size of 0.8 and testing sets with the size of 0.2 in our evaluation. The seeds and partition indices can be found on our GitHub repository.

\textbf{Existing alternatives.} We compare our E3C model with the following state-of-the-art alternatives.

$\bullet$
\textbf{JEI-DNN} is a state-of-the-art dynamic neural network model with early-exit branches, which enhances the backbone Transformer model with jointly optimized lightweight and trainable gates and intermediate classifier modules~\cite{regol2023jointly}. 

$\bullet$
\textbf{GFNet} is a conceptually simple yet computationally efficient model, which learns long-term spatial dependencies in the frequency domain with log-linear complexity~\cite{rao2023gfnet}. We have more detailed introduction in Section~\ref{S:distillation}.

$\bullet$
\textbf{ViT} processes images as sequences of patches using Transformer-based self-attention, and serves as the backbone model for many variants~\cite{dosovitskiy2020image}. In this paper, we choose the TinyViT variant, in which an efficient small vision transformer is obtained by transferring knowledge from a large pretrained model~\cite{wu2022tinyvit}. 

\subsection{Performance Evaluation}
The performance of E3C will be compared with the aforementioned alternatives, followed by ablation study and sensitivity test.

\subsubsection{Model comparison}
Table~\ref{tab:ablation} shows the comparison of our E3C model with state-of-the-art DNN models including GFNet, TinyVit, JEI-DNN in terms of model size, classification accuracy, and inference energy consumption. For each dataset in Table~\ref{tab:ablation}, the first row contains results from GFNet, the second row is from TinyViT, the third row is from JEI-DNN, and the last one (with distillation and early-exit both checked) is from E3C.

\textbf{Comparison of model accuracy.}
Similar to other lightweight DNN models, our E3C model suffers from accuracy loss due to the knowledge distillation and early-exit mechanisms.
However, E3C achieves over 94.7\% accuracy across four datasets, with accuracy loss less than 1.7\%, compared with the state-of-the-art baselines, as shown in Table~\ref{tab:ablation}. For real-time RSSC applications, E3C may decrease model accuracy slightly, but can significantly improve inference latency, model size, and energy consumption, as we discuss next. 
In addition, the E3C framework is backbone-agnostic, allowing it to be applied to more advanced DNN models beyond GFNet, potentially achieving even higher accuracy in the future.

\textbf{Comparison of DNN model size.}
Since we use different datasets to train our distillation and early-exit models, we report the model size, i.e., the number of model parameters of the E3C model as well as baseline models, for each and every dataset, as shown in Table~\ref{tab:ablation}. 
We see that the model sizes of E3C are all below 5MB across four datasets, significantly lower than the GFNet and JEI-DNN models, which have similar model sizes, since no distillation is applied. On average, E3C achieves a 71.8\% reduction in model size compared to GFNet, making it highly suitable for deploying RSSC models on resource constrained edge devices. 
We also see that the E3C model has larger mode sizes than the TinyViT model with distillation applied to the ViT model. This is due to the fact that E3C has additional early-exit network modules, e.g., the IMs and GMs modules. Note that the model size of E3C is only slightly larger than that of TinyViT, e.g., 4.57MB vs. 4.55MB for the UCM dataset. However, a slight increase in model size leads to a significant improvement in energy efficiency, as we discuss next. 

\textbf{Comparison of energy consumption.}
For a given dataset, the same RSSC model exhibits varying energy consumption across different edge devices. Thus, we show the energy consumptions from the E3C and baseline models on three different edge devices in Table~\ref{tab:ablation}. GFNet shows the highest inference energy consumption among the RSSC models, as it employs neither distillation nor early-exit mechanisms. The E3C model demonstrates significant energy savings on the Jetson AGX Orin device, achieving an average reduction in energy consumption of 47.2\% across four datasets compared to the GFNet model. Similarly, E3C achieves an average energy saving of 56.6\% on the Jetson Orin Nano device, compared to GFNet. For the Raspberry Pi edge device, E3C achieves an average energy saving of 66.9\%, compared to GFNet. Compared to JEI-DNN, E3C achieves an average energy consumption reduction of 41.3\% to 45.5\% on two heterogeneous edge devices, and a 48.7\% reduction on the CPU-only Raspberry Pi device. Thus, based on the results in Table~\ref{tab:ablation}, we see that the energy savings achieved by E3C stem from its early-exit module, which enables input images to terminate inference early, as well as from the distillation mechanism, since the smaller model size helps reduce both computational cost and redundancy. More detailed comparison of the energy efficiency performance of the E3C and other RSSC models are available in the supplementary material. 

\textbf{Comparison of model inference latency.}
For the real-time RSSC application on edge devices, DNN model inference latency is also an important performance metrics. Fig.~\ref{fig:latency} shows the inference latency results for E3C and baseline models. From Fig.~\ref{fig:latency}(a), we see that E3C achieves up to 1.7x inference speedup compared to the JEI-DNN model on the Jetson AGX Orin device. However, the GFNet model has slightly better performance than E3C for the UCM, NWPU and NaSC datasets on Jetson AGX Orin. This is primarily because the Jetson AGX Orin features high-performance CPU-GPU modules, whereas the Jetson Orin Nano and Raspberry Pi have limited computational resources. As a result, the early-exit mechanism offers limited benefit for high-performance processors. We further analyze the effects of early-exit on different datasets in Section~\ref{S:sensitivity}. As shown in Fig.~\ref{fig:latency}(b), on the Jetson Orin Nano device, E3C demonstrates superior inference latency performance compared to alternatives. Compared to GFNet, E3C achieves up to a 1.6x inference speedup on average. Finally, for Raspberry Pi with only CPUs, the inference latencies of all models are significantly higher than those from Jetson AGX Orin and Orin Nano. This is expected since Raspberry Pi has the most limited computational resources among the three edge devices. However, the model inference speedup from our E3C model is also the most significant, compared with the other edge devices. For example, the inference latency from E3C is below 450 milliseconds, while the latencies from alternatives are all above 750 milliseconds for the UCM dataset. Compared to GFNet, E3C achieves up to 3.1× speedup and compared to JEI-DNN, E3C achieves up to 1.9× speedup on resource constrained devices. To summarize, while the inference latency of GFNet is comparable to that of E3C on the NWPU dataset, our E3C model achieves 1.3× speedup compared to the state-of-the-art JEI-DNN model on average, and significantly outperforms all alternative models including GFNet, JEI-DNN and ViT on Jetson Orin Nano and Raspberry Pi edge devices for all the other datasets. 

\begin{figure*}
    \centering
    \includegraphics[width=0.95\linewidth]{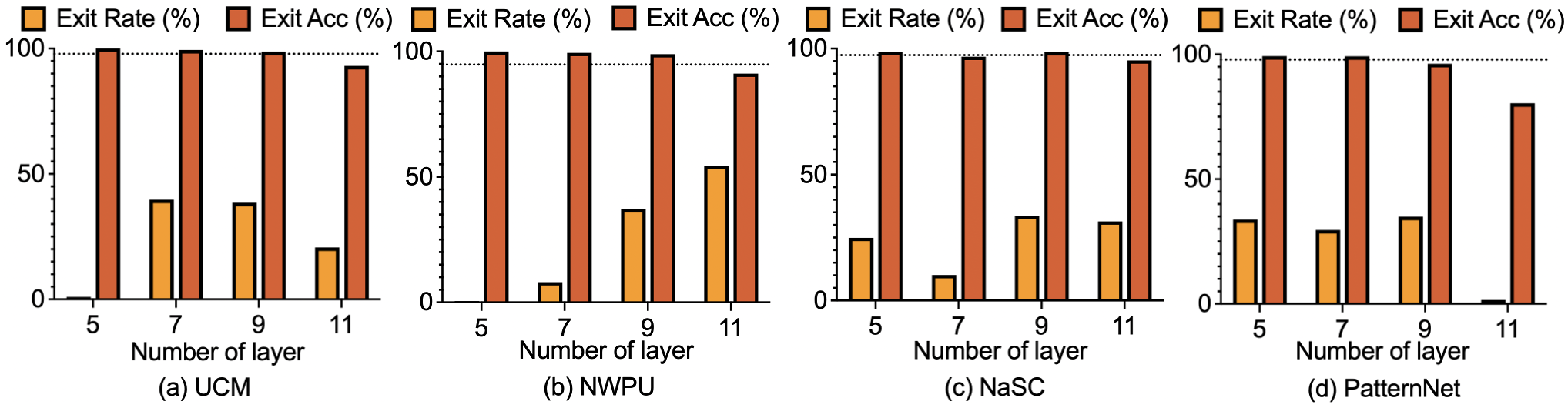}
    \caption{Exit rate and exit accuracy of E3C in four different datasets. The dotted line represents the overall model accuracy.}
    \label{fig:four_datasets}
\end{figure*}

\subsubsection{Ablation Study} \label{S:ablation}
We perform ablation study and investigate the effects of different parameters on the E3C performance. 

\textbf{Impacts of distillation and early-exit modules.} 
We conduct ablation experiments to study the impacts of frequency domain distillation and early exit on model size, accuracy and energy efficiency. 
From Table~\ref{tab:ablation} we see that the reduction in model size mainly comes from the knowledge distillation module. For example, the number of parameters of the distilled GFNet model is 4.57MB for the UCM dataset, a 71.4\% reduction from 15.92MB of the original GFNet model. 
In terms of model accuracy, the early-exit module does not cause much degradation, with the accuracy loss within 0.3\%, while the knowledge distillation module keeps the accuracy loss within 1.4\% across all four datasets. 

In addition, we see that the reduction in energy consumption during model inference comes from both distillation and early-exit. For example, knowledge distillation brings energy consumption on Jetson AGX Orin down from 144.8mJ to 75.3mJ, a 48\% reduction for the UCM dataset, while early-exit reduces energy consumption by 13\%. Adding early-exit modules increases model sizes slightly, but it can further reduce energy consumption, without sacrificing much accuracy. The energy consumption reduction rate for early exit ranges from 10.4\% to 37.8\%, while it ranges from 27.5\% to 63.7\% for distillation. 
Although adding distillation has more improvement on energy efficiency than adding early-exit modules on the UCM, NWPU and PatternNet datasets, the model accuracies of DNN models with early-exit modules are higher than those of models with distillation modules across all four datasets.
Compared with the original GFNet model without distillation, E3C has the most significant performance improvement, achieving energy savings of 30.2\% to 75.3\%. It reveals the effective complementarity of frequency domain distillation and early-exit methods in reducing inference energy consumption.
Note that the Raspberry Pi edge device has the fewest computational resources among three edge devices, and it consumes more energy running RSSC models. For the same UCM dataset, our E3C model running on Raspberry Pi consumes 432.2 mJ, while it only consumes 40.7 mJ on Jetson Orin Nano. However, it is on the resource constrained Raspberry Pi device that E3C has the most significant improvement on energy efficiency, with over 60\% improvement upon the original GFNet model across all datasets. 

Another interesting observation is that for the NaSC dataset, the DNN model with early-exit outperforms the DNN model with distillation in terms of both model accuracy and energy efficiency on Jetson AGX Orin. 
Since the NaSC dataset primarily comprises of ``easy'' imagery data, the DNN model with early-exit can make a significant portion of input images exit at shallow layers, thus significantly reducing inference latency and energy consumption, while maintaining high accuracy. Our data analysis shows that a significant amount of input images can exit within the first few early-exit branches for the NaSC dataset. We analyze the impacts of exit rate and datasets next. 

\textbf{Impacts of datasets and exit rate.}
As mentioned above, the early-exit model shows different performance on different datasets, since certain data inputs may not be ``easy enough'' to be determined by early classifiers. For different datasets, Fig.~\ref{fig:four_datasets} shows the exit rate and corresponding exit accuracy statistics defined in Section~\ref{S:settings}. 
We see that a significant amount of input images exit in the range of 6th to 11th layers.
For example, for the UCM and NWPU datasets shown in Fig.~\ref{fig:four_datasets}(a) and (b), almost no samples exit at the $5$th layer of the E3C model. If we add early-exit modules to each and every layer of the backbone model as in the JEI-DNN model, it will increase computational redundancy and cause relatively high inference latency. 
However, for the NaSC dataset statistics shown in Fig.~\ref{fig:four_datasets}(c) and our supplementary material, more input imagery data can exit from the 5th layer or even earlier layers with high exit accuracies. It explains the reason why the early-exit models behave differently for different datasets.

\subsubsection{Sensitivity Test} \label{S:sensitivity}
We perform sensitivity test on the early-exit model parameters next. 

\begin{figure}[tbp]
\centering
\includegraphics[width=0.95\linewidth]{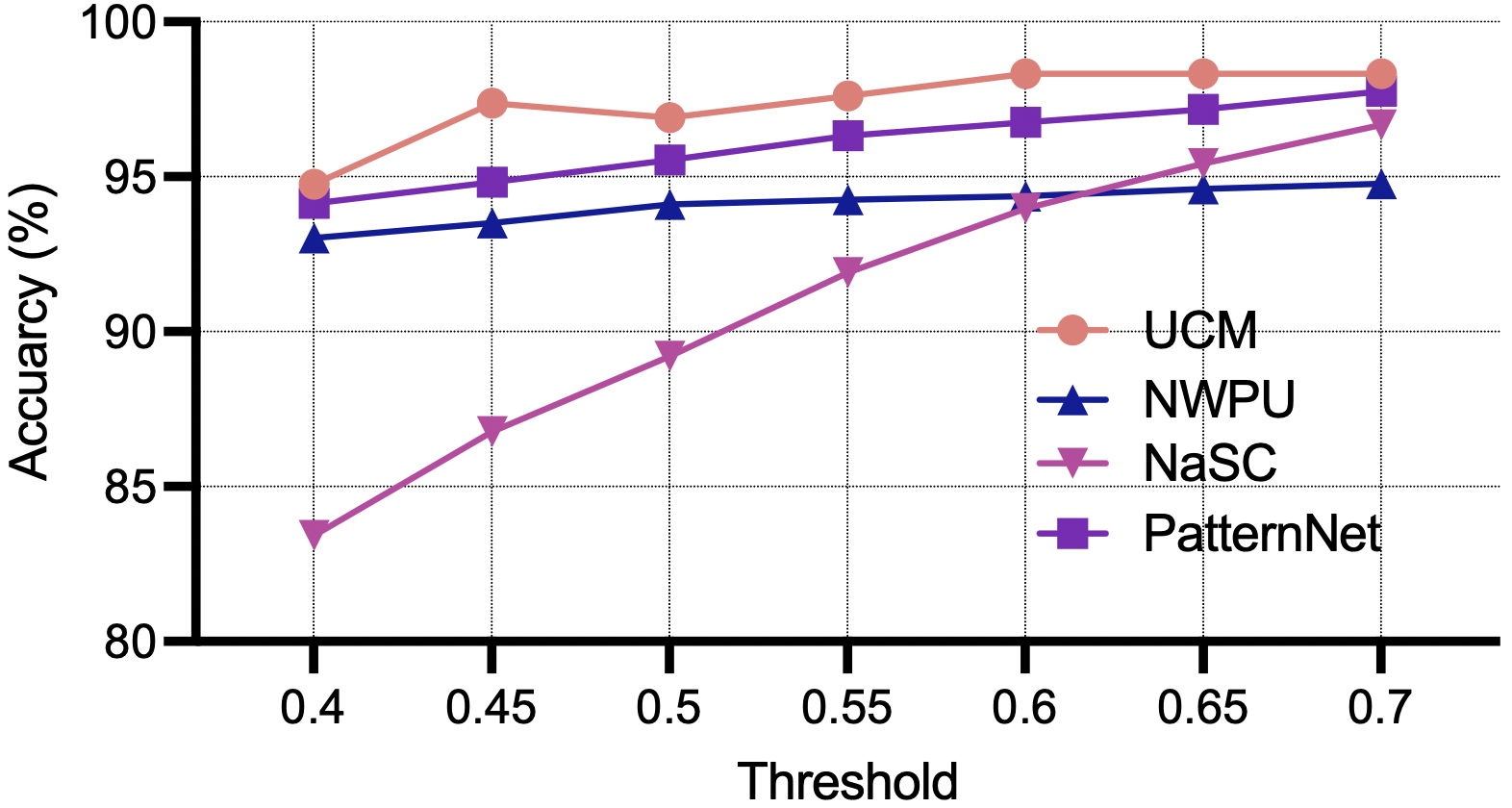}
\caption{The classification accuracy of E3C under different thresholds across four datasets.}
\label{fig:threshold_accuracy}
\end{figure}

\textbf{Impact of the early-exit threshold parameter.}
As mentioned in Section~\ref{S:early-exit}, a threshold parameter is used in the gating mechanism of the early-exit module. 
As shown in Fig.~\ref{fig:threshold_accuracy}, we report the accuracy of E3C under various threshold values. We see that as the threshold value increases, the model accuracy also increases in general. It makes sense since the larger the threshold value is, the more strict the conditions are for early exit, which can lead to higher accuracy with higher computational cost. In addition, for most datasets, the model accuracy is not sensitive to the threshold value in a wide range from 0.4 to 0.7. For the NaSC dataset, since more input imagery data can exit from early layers with lower accuracy, as shown in the supplementary material, increasing the threshold value can move the exit points to later layers, thus leading to higher accuracy. However, for threshold greater than 0.7, model accuracy has diminishing returns for all datasets. 

\begin{figure}[tbp]
\centering
\includegraphics[scale=0.33]{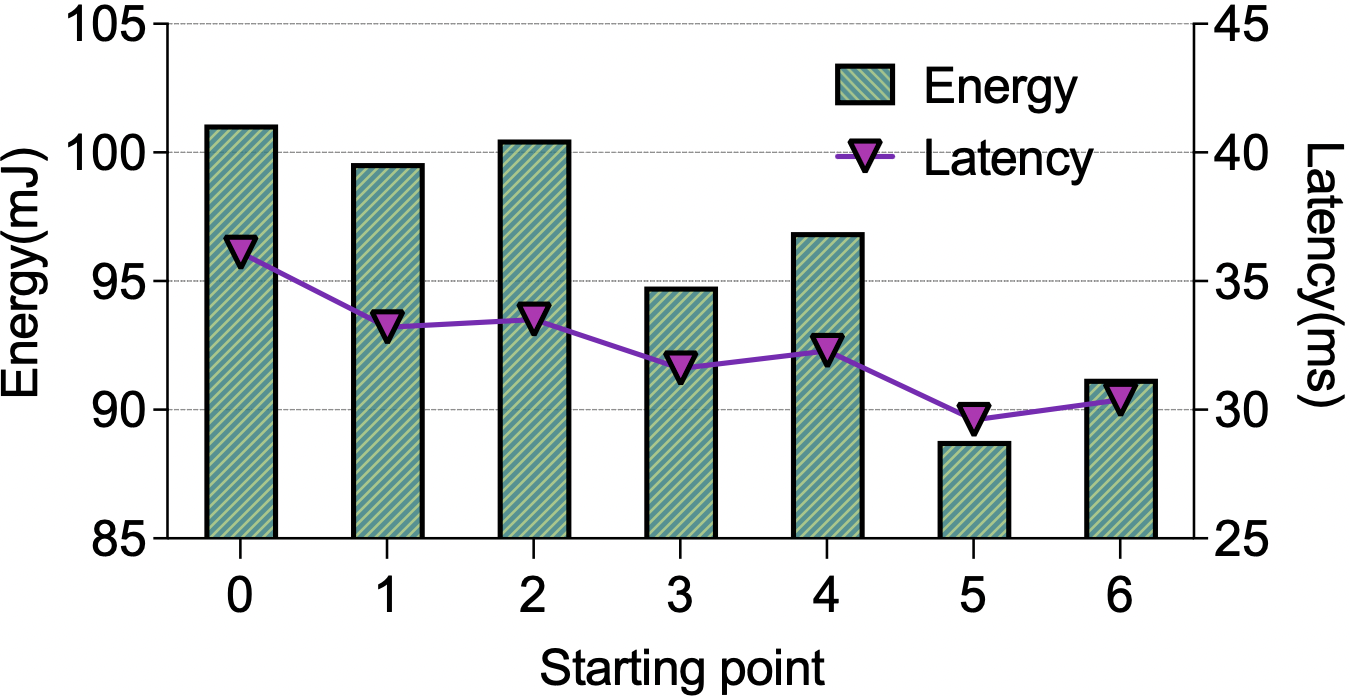}
\caption{Comparison of inference energy consumption and inference latency for different starting points using the NWPU dataset on Jetson AGX Orin.}
\label{fig:start_point}
\end{figure}

\textbf{Impact of the early-exit starting point parameter.}
As mentioned in Section~\ref{S:lightweight-early-exit}, we start incorporating early-exit branches from an intermediate layer of an index of $l_m$ to reduce the early-exit computational overhead. 
Based on our statistic analysis of exit rate in Section~\ref{S:ablation}, we find that for our backbone model GFNet with 12 layers, a significant amount of data exit after the 5th layer with high classification accuracy, as shown in Fig.~\ref{fig:four_datasets}. Thus, we set the model parameter $l_m=4$ based on the statistics. Here, we show the effect of this parameter on the model latency and energy consumption. 
We set $l_m$ in the range of 0 to 6 representing the first seven layers, as the starting point of adding early-exit branches. We run our E3C model using the NWPU dataset on the Jetson AGX Orin edge device, and measure the model inference latency and energy consumption for different starting points. As shown in Fig.~\ref{fig:start_point}, both energy consumption and latency reach their minimum values when $l_m=4$, which is consistent with our statistic analysis. In fact, when we set the parameter $l_m$ too low, the model may struggle to capture the features of certain input images with shallow layers, producing high inference latency and energy consumption.
If the parameter $l_m$ is set too high, the early-exit branches will be shifted too far back, without much benefit of early-exit. Thus, this parameter is a trade-off between model accuracy and efficiency, which should be determined from training data statistics, as discussed in Section~\ref{S:settings}.  

\section{Conclusion}
This work investigates the complementary benefits of the knowledge distillation and early-exit methods for lightweight RSSC DNN models on edge devices. 
We first perform frequency domain distillation on the GFNet model, and then design a new early-exit model and integrate it with the distilled GFNet model.
For edge devices with CPUs and GPUs, we distribute the compute-intensive task of early-exit model parameter computation to CPUs, while keeping other model parameters on GPUs to better utilize heterogeneous computing resources.
Extensive experiments with four RSSC datasets on three edge devices show that our E3C model outperforms state-of-the-art alternatives in terms of model size, inference latency, and energy efficiency, without sacrificing much model accuracy. 
Finally, the E3C framework is backbone-agnostic. Thus, this work can pave the way for further study of applying lightweight DNN models to other applications on edge devices. 


\bibliographystyle{ACM-Reference-Format}
\balance
\bibliography{sample-base}

\end{document}